\documentclass{article}
\usepackage{graphicx} 
\usepackage[hyphens]{url}
\usepackage{authblk}
\usepackage[natbib,
backend=biber,
style=authoryear,
sorting=ynt
]{biblatex}
\title{Response: Emergent analogical reasoning in large language models}
\author[1]{Damian Hodel}
\author[1]{Jevin West}
\affil[1]{Center for an Informed Public, Information School, University of Washington}
\affil[ ]{\textit{hodeld@uw.edu, jevinw@uw.edu}}
\date{\today}

\begin{document}

\maketitle

\section{Introduction}

In their recent Nature Human Behaviour paper, ''Emergent analogical reasoning in large language models,'' \citep{webb_emergent_2023} the authors argue that ``GPT-3 exhibits a very general capacity to identify and generalize—in zero-shot fashion—relational patterns found within both formal problems and meaningful texts.'' This conclusion arises from their comparison of GPT-3 with human performance across four analogical reasoning domains, where they find comparable results. In this response, we argue that this approach is unsuitable for evaluating \textbf{general, zero-shot reasoning} in large language models (LLMs). Two primary reasons underlie our objection. 
First, the term ``zero-shot'' implies problem sets entirely novel to GPT-3. However, the chosen approach cannot conclusively eliminate the possibility of these problems residing in the LLM's training data, as acknowledged by the authors themselves in the review file\footnote{\url{https://www.nature.com/articles/s41562-023-01659-w#peer-review}}. Second, the assumption underlying this approach is that tests designed for humans can accurately measure LLM capabilities. This assumption is prevalent, but remains unverified. 
We also provide empirical results to support our claims, see appendix (Section~\ref{sec:counterexamples}). 
Our counterexamples show that GPT-3 fails to solve simplest variations of the original tasks, whereas human performance remains consistently high across all modified versions.

Given the hype surrounding LLM capability and this paper in particular\footnote{\url{https://www.tagesanzeiger.ch/beherrscht-die-kuenstliche-intelligenz-analogien-768020720454}}\footnote{\url{https://www.news-medical.net/news/20230731/AI-language-model-GPT-3-performs-about-as-well-as-college-undergraduates-in-analogical-reasoning.aspx}}\footnote{\url{https://www.sciencemediacentre.org/expert-reaction-to-study-looking-at-gpt-3-large-language-model-and-ability-to-reason-by-analogy/}} contrasted by the many findings of LLM brittleness\footnote{\url{https://www.technologyreview.com/2023/08/30/1078670/large-language-models-arent-people-lets-stop-testing-them-like-they-were}}, we felt it was important to respond and illustrate the insufficiency of the methods employed in addressing GPT-3's supposed general, zero-shot reasoning. It is important that we interpret LLM results with caution and refrain from LLM anthropomorphization. Tests designed to assess the general capabilities of humans may not inherently serve the same purpose when applied to LLMs.

Others have commented on this paper and we want to note these contributions. \citet{Mitchell_2023} discusses this paper, focusing on the letter string and digit matrix analogy problems. Mitchell disagrees that "the digit matrix problems are essentially equivalent in complexity and difficulty to Ravens Progressive Matrix problems." Further, Mitchell presents individual counterexamples of the letter string problems where GPT-3 makes nonhuman-like errors, as evidence against the claimed robustness of GPT-3 in analogy reasoning. We conduct a similar but more systematic analysis and include human behavioral experiments, as detailed in the appendix, that concurs with Mitchell's conclusion. Mitchell also points out that the term "accuracy" implies that there was only one correct answer to each problem, which isn't the case with these problems, but an assumption implicitly made by the authors. For comparison purposes, we adopt \citet{webb_emergent_2023}'s assumption in our paper and use the same terms, i.e. ``accuracy'' and ``performance'' but recognize this limitation.

\section{Criticism of the Methods Employed in the Original Paper}
\label{sec:criticism_methods}
To assess general, zero-shot reasoning capacity of LLMs, \citet{webb_emergent_2023} compare GPT-3 with humans and find similar or even better performance across a range of analogical reasoning tests adapted from existing cognitive tests designed for humans. 
However, we believe that this approach is not sufficient for testing the general, zero-shot reasoning capacity of large language models (LLMs). Here is why: \\

First, ``zero-shot'' implies analogical problem sets that are entirely novel to GPT-3, encompassing both specific examples and variants of those examples. However, this condition is not met by some of the letter string problems used in the original paper, as noted by the authors themselves in the review file\footnote{\url{https://www.nature.com/articles/s41562-023-01659-w#peer-review}}: ``It is possible that GPT-3 has been trained on other letter string analogy problems, as these problems are discussed on a number of webpages.'' Without ruling out the possibility of data memorization, one cannot claim zero-shot reasoning. As the first author notes in a recent MIT Technology Review article\footnote{\url{https://www.technologyreview.com/2023/08/30/1078670/large-language-models-arent-people-lets-stop-testing-them-like-they-were}}, [if the test examples exist in the training data], ``I think we really can’t conclude much of anything.''

In the before-mentioned peer review file, the authors further note that they ask GPT-3 about these problems as a way of testing their existence in the training data. It makes sense to at least try this, but we find this to be weak evidence, given the large number of possible answers to this question and the ambiguity of the answers given. To investigate this further, we asked ChatGPT to provide examples of letter string problems. Examples were given, suggesting that it has seen such examples in the training data. We include our question and ChatGPT's answer in the appendix. Important to note is that ChatGPT was trained on more data than GPT-3 so this result only provides circumstantial evidence.  

Zero-shot reasoning is an extraordinary claim that requires extraordinary evidence. At the very least, it necessitates demonstrating that the problems, as well as their variations, do not already exist within the training data, as previously mentioned. The original paper fails to offer such evidence for any of the four task domains. We do recognize that obtaining such evidence can be exceptionally challenging. Many researchers lack access to GPT-3's training data, and even if they did, confirming the absence of examples or derivations from the training data is nearly impossible. However, the difficulty to provide evidence of zero-shot should not be a reason to claim it.   \\

Second, \citet{webb_emergent_2023} claim that the presented problem types test GPT-3's human-like reasoning capacity in a "very general" way. This assumption is based on the premise that LLMs behave similarly to humans, thus implying that a test designed for humans can adequately assess LLMs in a broader capacity beyond the tasks included in the test. However, this assumption has not been substantiated. 

On the contrary, generalized findings across the literature of LLM brittleness tend to contradict it. In Appendix~\ref{sec:counterexamples}, we present counterexamples involving the letter string analogy problems, which demonstrate the brittleness of the assessment approach employed. In these tests, GPT-3 fails to solve simple variants of the letter string analogies presented in the original paper, while human performance remains on a high level. 

In addition to the finding that GPT-3 matches or even outperforms human performance, \citet{webb_emergent_2023} further show that GPT-3 exhibits human-like characteristics in analogical reasoning, i.e., decreasing performance with increasing problem complexity. Based on this result, the authors propose that GPT-3 may have developed mechanisms similar to those underlying human intelligence. This is one possible interpretation. However, an alternative explanation could be that the training data contains a scarcity of solutions to complex problems, possibly reflecting the challenges humans encounter with such problems, a notion supported by our experiments involving human subjects. 

It is important to note that our intention is not to discredit the use of such tests for studying LLMs but to point out the limitations of these methods for making claims about the reasoning capacity of LLMs. \\

Before conducting the human behavior experiments, we shared our counterexamples on GPT-3 with the first author of the original paper, and greatly appreciate their engagement in this discussion. One of their main objections was the expectation that our modified problems would also be significantly more difficult for human subjects. The human behavioral studies we carried out definitively contradict the predictions of the primary author. Despite a notable decrease in GPT-3's performance on our adapted tasks, humans consistently demonstrate strong performance. Nevertheless, it is important to note that comparing performance to humans, whether better or worse, is not evidence of the claimed capacity. For example, if one ran this comparison only among humans and two groups emerged from the sampling, one with adults and one with children\footnote{This analogy presumes that LLMs' performances can be assessed against human standards. It is important to clarify that we don't endorse this assumption until it is substantiated, but for the sake of our argument, we will adopt it from the original paper.}, we would likely find that adults outperform children on these reasoning tasks. According to the authors' logic \citep{webb_emergent_2023}, this would be evidence against zero-shot reasoning in children. But we know that children have this ability. Hence, performance compared to humans cannot be used to support or refute zero-shot reasoning.

\section{Conclusion} 
Based on their analysis, \citet{webb_emergent_2023} argue that LLMs have acquired a general ability for zero-shot reasoning. With full respect to the authors and their work, we disagree with this interpretation. As we show and argue in our response, the methods are insufficient to evaluate a capacity for true, zero-shot reasoning. 
Given the current hype surrounding LLMs, we hope this can be used to spur further tests and evaluations of what LLMs can and cannot do.

\section{Code and data availability}
Code and data can be downloaded from: \url{https://github.com/hodeld/emergent_analogies_LLM_fork}

\printbibliography

\section{Author contributions}
D.H. conducted the experiments. D.H. and J.W. drafted the manuscript.

\section{Competing interests}
The authors declare no competing interests.

\newpage

\section{Appendix}

\subsection{Counterexamples}
\label{sec:counterexamples}
To investigate whether the problems presented in the original paper truly assess analogical reasoning in GPT-3 or primarily its capability to recite training data, we create non-standard variants of the original tasks that are less likely to be found in training data. Our focus is on the letter string analogies, a subset of the four problem domains examined, and we conduct tests with both human subjects and GPT-3. In our experiments, GPT-3 performance significantly declines when presented with these additional counterexamples, while human performance remains consistently high across all tests (\ref{fig:gpt3_human_synthetic_alphabet}). This suggests that the claims made in the original paper regarding GPT-3's zero-shot reasoning may not be substantiated.

\subsubsection{Methods}

\begin{figure}[ht]

  \centering
  \includegraphics[width=0.75\linewidth]{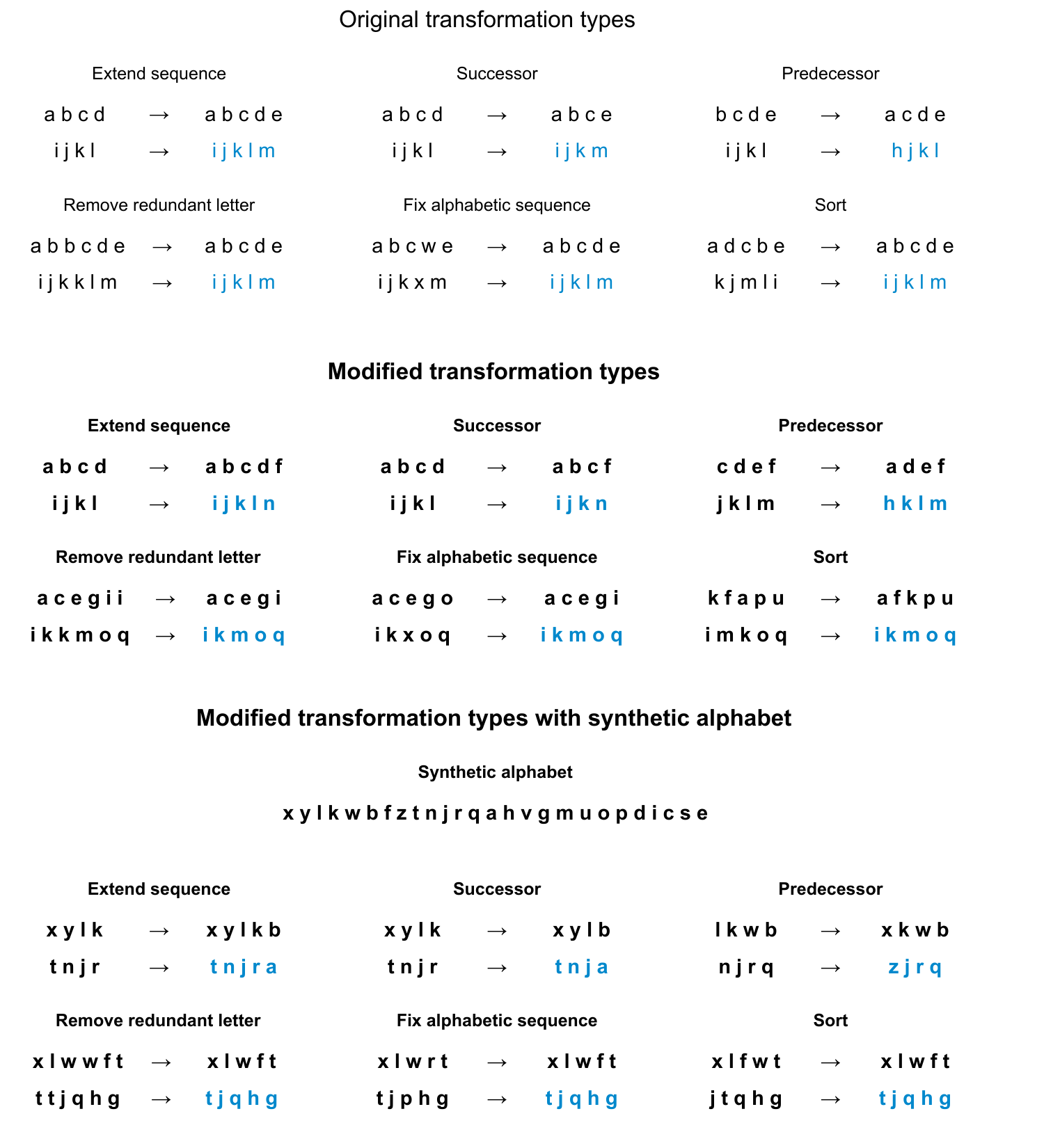}
  \vspace{-3mm}
  \caption{\small Letter string analogies along their transformations of both the original paper and our counterexamples. We introduce a synthetic alphabet into the task and apply two types of letter sequence modifications, both based on increasing the interval from one to two letters. For the transformation types 'extend sequence', 'successor', and 'predecessor', the modification only affects the \textit{letter to change} (last or first letter). For 'remove redundant letter', 'fix alphabetic sequence', and 'sort', the interval is increased for the complete letter sequence. We apply the same modifications to the problems generated with the synthetic alphabet. }
  \label{fig:trans_types}
    \vspace{-3mm}
\end{figure}

In order to test GPT-3's generality in zero-shot analogical reasoning, we extend the letter string analogies with two modifications and compare GPT-3's and humans' performance analogous to the original approach. The modifications involve using a synthetic alphabet and increasing the size of the interval from one to two letters, see Figure~\ref{fig:trans_types}. If the claim regarding GPT-3's zero-shot reasoning capability is true, we can expect similar performance across modifications, in particular, independent of the alphabet. Unlike the original study, we view the comparison with human performance not as evidence for or against GPT-3's analogical reasoning abilities, but rather as a confirmation of the validity of our set of problems.

We create the synthetic alphabet by randomly changing the order of the letters in the real alphabet.  
For both humans and GPT-3, we incorporate the synthetic alphabet in the tasks by preceding the original prompt with the sentence ``Use this fictional alphabet: \verb|[|x y l k w b f z t n j r q a h v g m u o p d i c s e\verb|]|.''

The increase in the size of the interval from one to two letters aims to rule out the possibility that GPT-3 merely replicates the fed sequence of letters. We achieve this in two ways. For the problem types 'extend sequence', 'successor', and 'predecessor', we increase the interval size for the \textit{letter to change} from one to two. For the problem types 'remove redundant letter', 'fix alphabetic sequence', and 'sort', we increase the interval size of the \textit{complete letter sequence} from one to two \footnote{It is worth noting that we apply this modification to both the source (the first row for each example in Figure~\ref{fig:trans_types}) and the target (the second row for each example in Figure~\ref{fig:trans_types}), minimizing the difficulty of the modified problems and allowing us to compare our tests to the zero-generalization problems given in the original paper.}. 

We compare GPT-3's and human performance for the following three settings: the original tasks as reported in \citep{webb_emergent_2023}, counterexamples that involve the interval size modification, and counterexamples that include both the interval size modification and the synthetic alphabet. To ensure that GPT-3 is capable of processing the introduced modifications \citep[``counterfactial comprehension check'']{wu_reasoning_2023}, we additionally include tests on GPT-3 for two additional settings: original examples on the real alphabet but including the modified prompt, i.e. ``Use this fictional alphabet: \verb|[|a b c d e f g h i j k l m n o p q r s t u v w x y z \verb|]|. ''), and counterexamples involving the synthetic alphabet but without increasing the interval size. 

\paragraph{GPT-3 evaluation}
Our code for reproducing Figure~\ref{fig:gpt3_human_synthetic_alphabet} is available on Github\footnote{\url{https://github.com/hodeld/emergent_analogies_LLM_fork}}. For each problem type, we create 50 instances to mirror the original paper. The settings are as follows: model variant=text-davinci-003, temperature=0, maximum length=20. 
Using the original code, we mirror the evaluation and analysis approach of the original paper.  The prompt pattern including the synthetic alphabet illustrates the following example.

\begin{small}
\verb|Use this fictional alphabet: [x y l k w b f z t n j r q a h|

\verb|v g m u o p d i c s e]. Let’s try to complete the pattern:|
\newline

\verb|[x y l k] [x y l k b]|
\newline

\verb|[t n j r] [|
\end{small}

\paragraph{Human behavioral experiment.}
We conducted human behavior experiments through an online study with University of Washington (UW) undergraduates analogous to the experiments of the original paper. All participants provided their informed consent prior to the study, and the data collection process was approved by the UW Institutional Review Board (IRB ID STUDY00019080, approved on 6 November 2023). 121 participants completed the study. They were compensated with extra course credits for their participation. 

The first author of the original study generously provided participant instructions, which we adapted for our experiments. In particular, we presented the participants an additional example problem to introduce the synthetic alphabet.  

\begin{center}
Use this fictional alphabet: \verb|[|x y l k w b f z t n j r q a h v g m u o p d i c s e\verb|]|.
\newline

\verb|[|x x x\verb|]| \verb|[|y y y\verb|]|\\
\verb|[|l l l \verb|]| \verb|[?|\verb|]|
\end{center}

Each participant completed a total of 18 zero-generalization tasks, consisting of six problems for each setting (one problem for each transformation type). 

\subsubsection{Results}

\begin{figure}[h]

  \centering
  \includegraphics[width=0.9\linewidth]{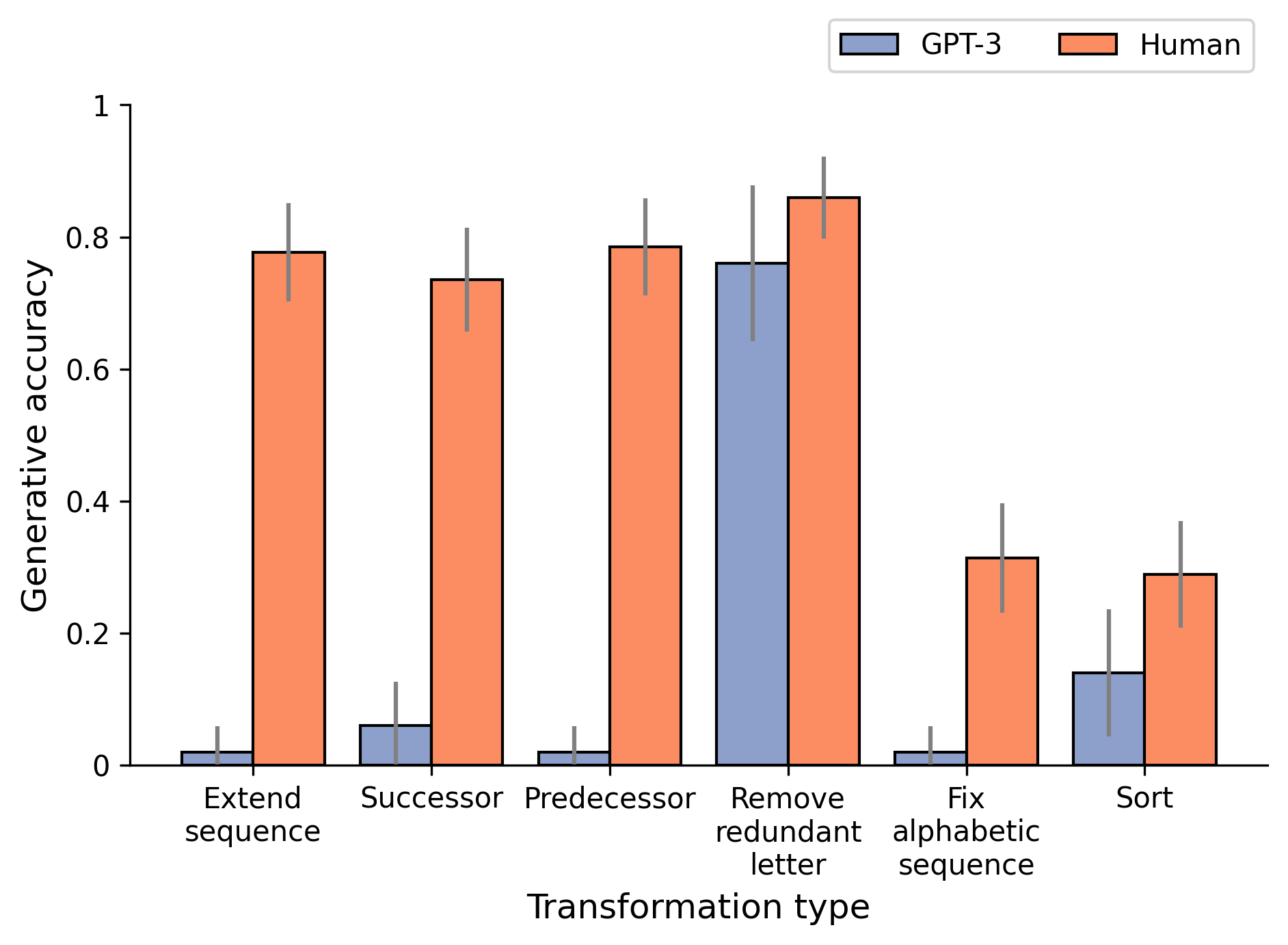}
  \vspace{-3mm}
  \caption{\small 
  Comparison between GPT-3's (blue) and human (orange) performances on modified letter string problems involving a synthetic alphabet and a larger interval size. The transformation types and their order correspond to Figure 6b in the original paper.
Humans demonstrate significantly higher accuracy compared to GPT-3. 
Human results represent the average performance of 121 participants (UW undergraduates). Each participant received one randomly selected instance of each problem subtype. GPT-3 results reflect the average performance across all 50 instances.
Gray error bars indicate 95\% binomial confidence intervals for the average performance across multiple problems.}
  \label{fig:gpt3_human_synthetic_alphabet}
    \vspace{-3mm}
\end{figure}

\begin{figure}[ht]

  \centering
  \includegraphics[width=0.9\linewidth]{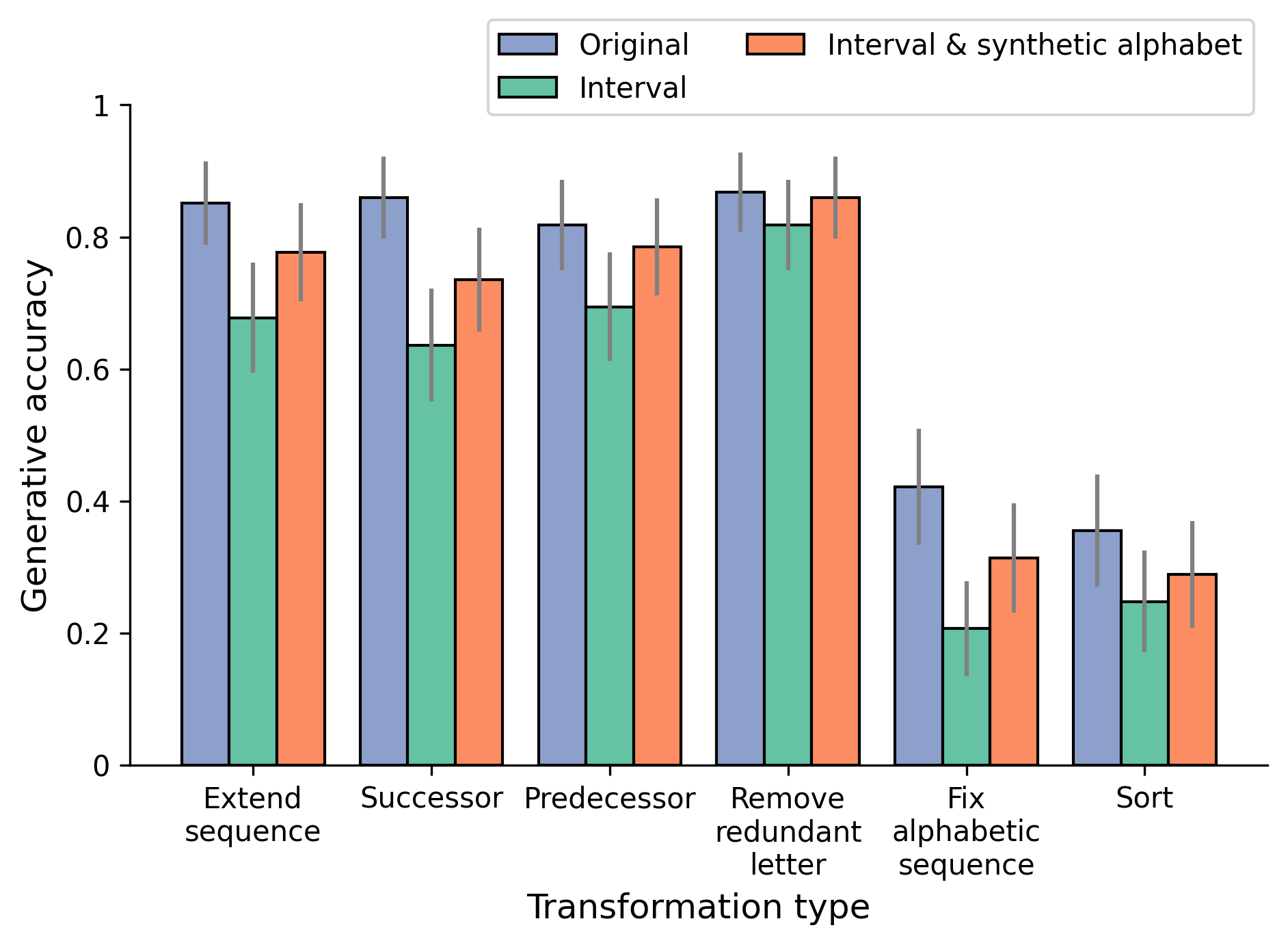}
  \vspace{-3mm}
  \caption{\small \textbf{Human} performance for zero-generalization letter string problems for the original experiment (blue) and with the larger interval size (green), and larger interval size with synthetic alphabet (orange). Human accuracy in the modified problems is comparable to that in the original problems (blue). The results reflect the average performance of N = 121 participants (UW undergraduates).}
  \label{fig:trans_comparison_humans}
    \vspace{-3mm}
\end{figure}

In our experiments, human achieved consistently higher accuracy than GPT-3, in particular on modified letter string tasks involving both the synthetic alphabet and increased letter interval size, see Figure~\ref{fig:gpt3_human_synthetic_alphabet}. Human performance remains at a level similar across modifications (Figure~\ref{fig:trans_comparison_humans}) while GPT-3 performance declines significantly for modified problem types (Figure~\ref{fig:trans_comparison_humans}). The generative accuracy of GPT-3 for the synthetic alphabet is close to zero ($<0.1$) when performing the modified tasks 'extend sequence', 'successor' or 'predecessor', and 'fix alphabetic sequence'. Only for 'remove redundant letter' and 'sort' does GPT-3 achieve accuracy in a range similar to that reported in the original paper \citep{webb_emergent_2023}. 

\begin{figure}[h]

  \centering
  \includegraphics[width=0.9\linewidth]{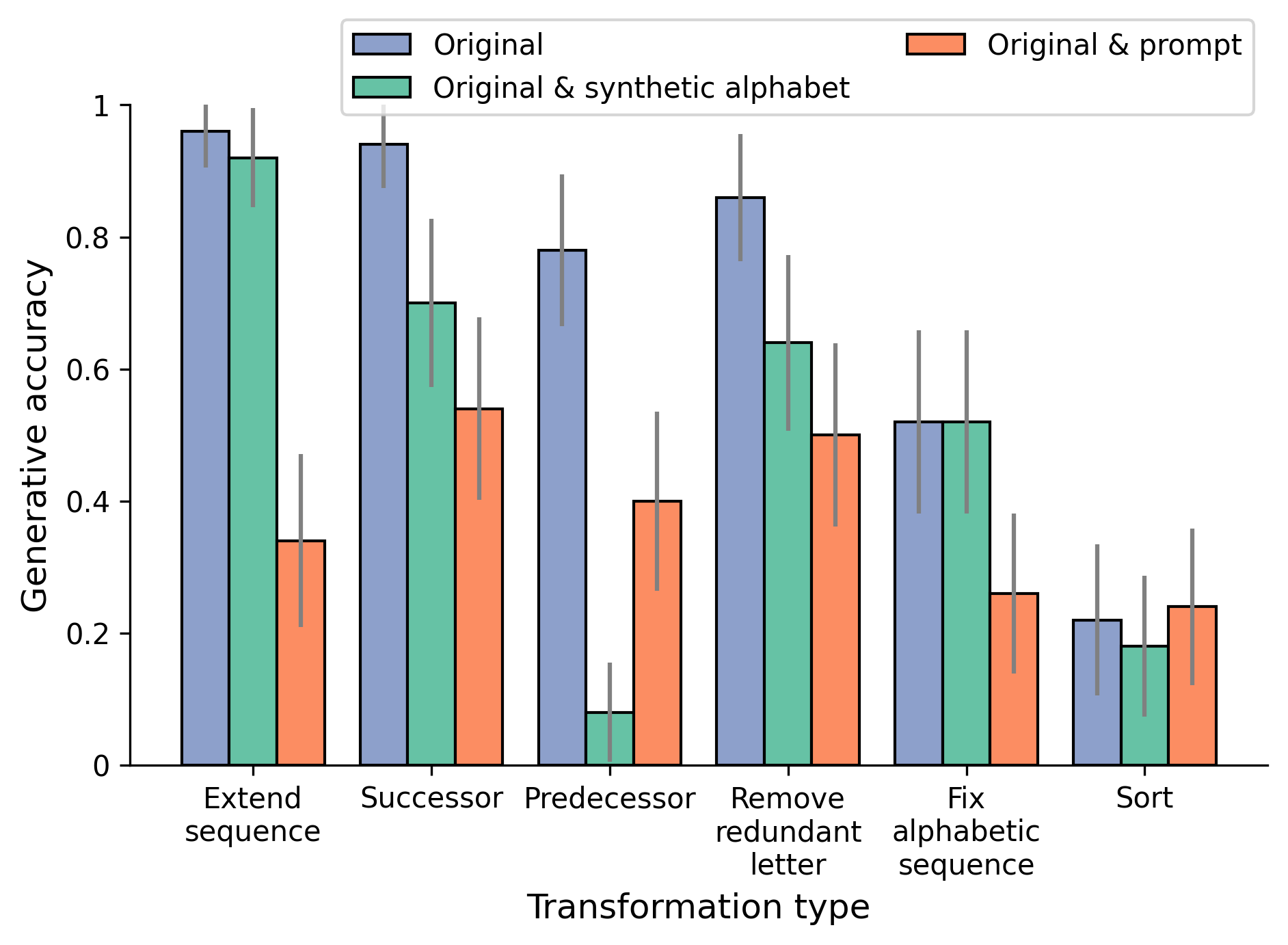}
  \vspace{-3mm}
  \caption{\small Counterfactual comprehension check. Comparison of GPT-3 performance on zero-generalization letter string problems between original tasks (blue) and the only marginally modified tasks involving a synthetic alphabet without modification of the interval size (green) and a modified prompt without modified string sequence (orange). The accuracy on modified tasks is lower than on the original ones but, greater than 0.2 except for `remove redundant letter' and `sort' involving the synthetic alphabet. The figure and the order of the transformation types correspond to Figure 6b in the original paper. These results reflect an average performance for N=50 instances.}
  \label{fig:comprehension_check}
    \vspace{-3mm}
\end{figure}

Figure~\ref{fig:comprehension_check} shows the accuracy of GPT-3 in the two counterfactual comprehension checks \citep{wu_reasoning_2023}. For all but on the 'precessor' task on the synthetic alphabet, we obtain a GPT-3 accuracy of at least 30\% of the original level, demonstrating GPT-3's ability to process the introduced modifications.

\begin{figure}[h]

  \centering
  \includegraphics[width=0.9\linewidth]{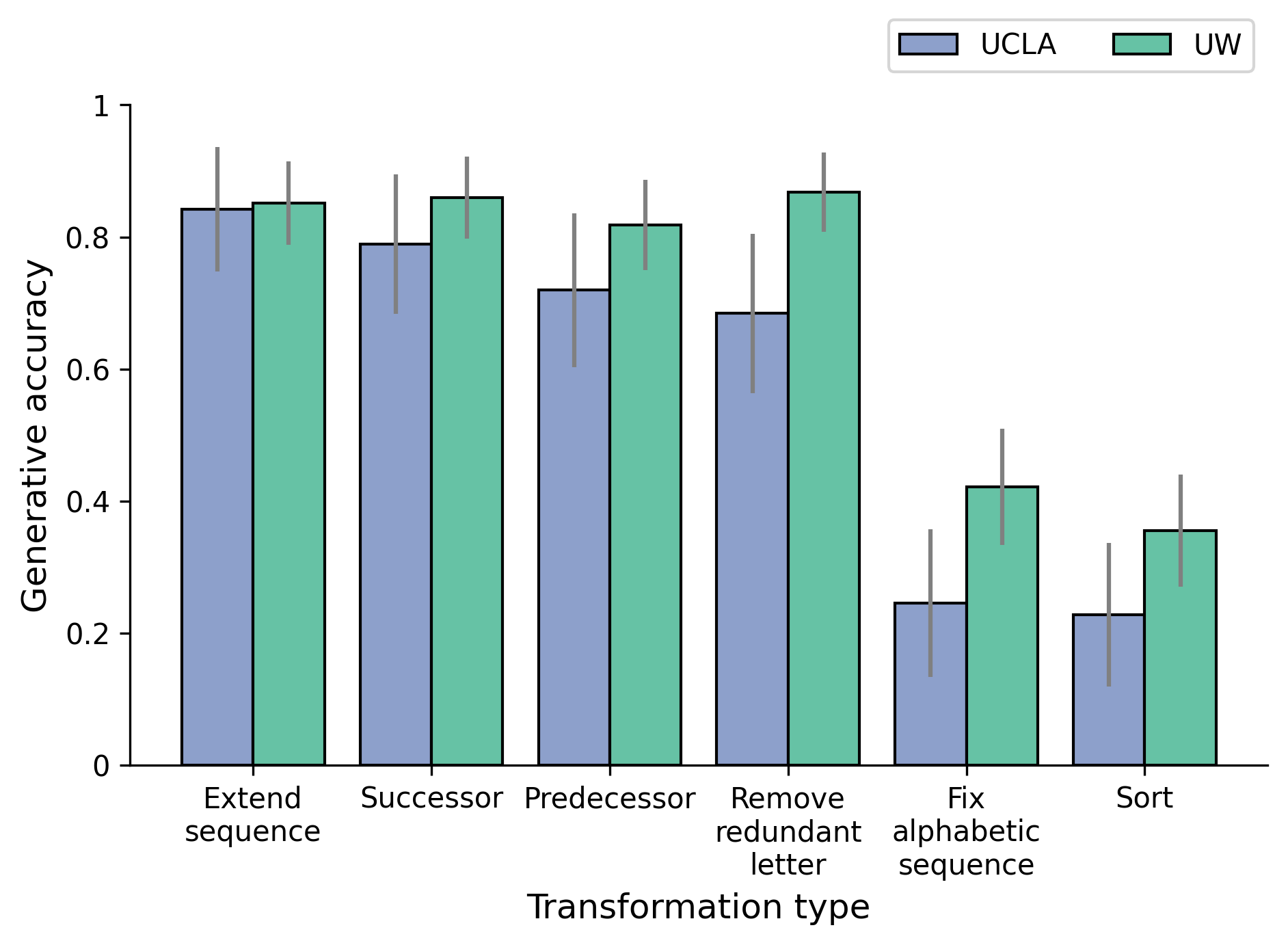}
  \vspace{-3mm}
  \caption{\small Comparison of human performance on the original letter string tasks between the outcomes reported in the original study (blue) and the findings presented in this paper (green). UW undergraduate students exhibit marginally higher accuracies. The transformation types and their order correspond to Figure 6b in the original paper.      }
  \label{fig:humans_ucla_uw}
    \vspace{-3mm}
\end{figure}

Lastly, Figure~\ref{fig:humans_ucla_uw} illustrates the comparison of human performance in the original tasks between the participants of the original study and those in our study. Although the subjects in our study marginally outperform those in the previous study, the similarity in performances is evidence that our experimental setup and execution align with the original study at UCLA.


\begin{figure}[ht]

  \centering
  \includegraphics[width=0.9\linewidth]{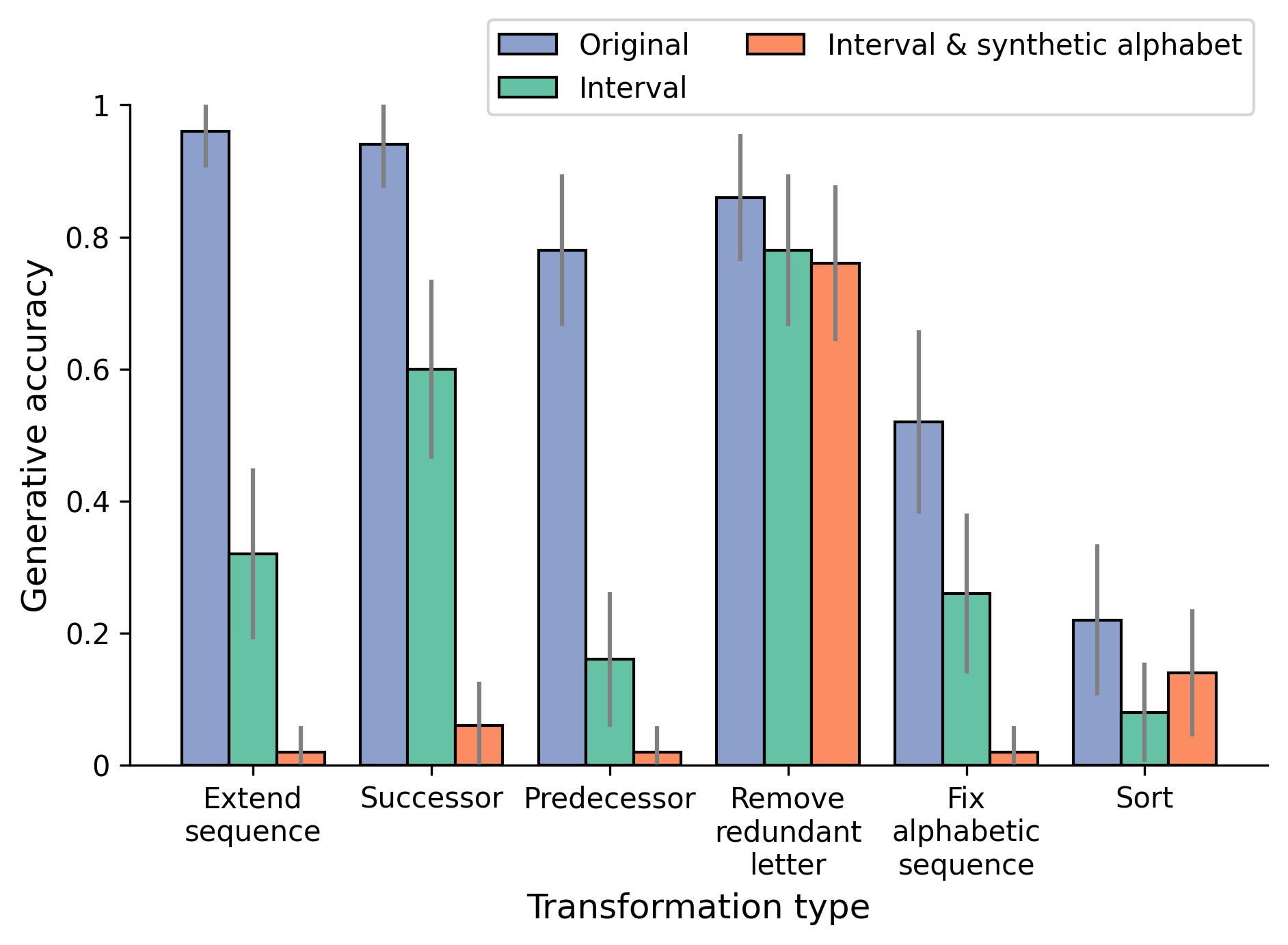}
  \vspace{-3mm}
  \caption{\small \textbf{GPT-3} performance for zero-generalization letter string problems for the original experiment (blue) and with the larger interval size (green), and larger interval size with synthetic alphabet (orange). Except for 'remove redundant letter,' GPT-3's accuracy declines significantly for the modified problems. The results reflect an average performance for N=50 instances.}
  \label{fig:trans_comparison_gpt3}
    \vspace{-3mm}
\end{figure}

\subsubsection{Discussion}
The recent paper, "Emergent analogical reasoning in large language models" \citep{webb_emergent_2023}, and subsequent news articles argue that LLMs may have acquired the emergent ability for zero-shot analogical reasoning. We are less certain of these conclusions, given our own follow-up experiments. Our results show low success of GPT-3 in solving letter string problems with simple modifications and with a synthetic alphabet, while human performance remains high. 

Only in two out of six problem types ('remove redundant letter' and 'sort'), GPT-3 achieves similar generative accuracy 
on our counterexamples compared to the original problems involving the real alphabet, as well as in comparison to human performance on the same modified problems. For these two problem subtypes, GPT-3 does not need to generate a letter from the full alphabet, but only to remove the duplicate letter or to rearrange given letters, which may explain the higher performance. 
The results of these two tasks also serve as an additional counterfactual comprehension check \citep{wu_reasoning_2023} in addition to the accuracy of GPT-3 under the only marginally modified conditions, shown in Figure~\ref{fig:comprehension_check}. The results demonstrate that GPT-3 is capable of processing synthetic alphabets, which validates our approach.

So what explains the high success of GPT-3 in solving the problems on the real alphabet (as used in the original paper) but failure with the synthetic alphabet and with the modified interval size for most of the letter string problems while human performance remains consistently high? 

Our results suggest that the answer resides in the training data confirming the analysis of the methods in Section~\ref{sec:criticism_methods}. Unlike humans, GPT-3 performs well only for simple analogy problems with the standard English alphabet, which are likely to be present in the training data. These findings contradict two of the main claims in the original paper \citep{webb_emergent_2023} regarding GPT-3's capacity for general, zero-shot reasoning and its human-like characteristics in analogical reasoning.
Consequently, we reject the proposition made in the original paper that GPT-3 may have developed mechanisms similar to those underlying human intelligence. 

The GPT-3 failure to solve simple variations of the original problems demonstrates the brittleness of the presented approach when assessing human-like reasoning in language models.


\subsection{ChatGPT's answer to our question: "Could you give an example of a copycat problem?"}
\begin{figure}[ht]

  \centering
  \includegraphics[width=0.9\linewidth]{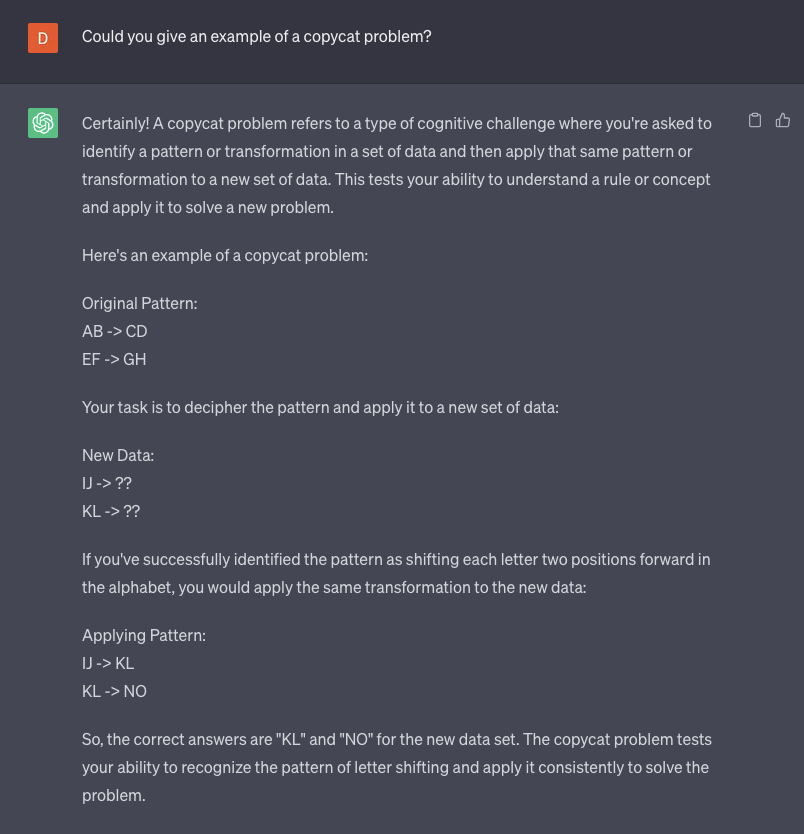}
  \vspace{-3mm}
  \caption{\small ChatGPT's answer to our question: "Could you give an example of a copycat problem?". The tasks presented in the original paper are called Copycat. This name refers to a computer program that tests such letter string analogy problems. }
  \label{fig:chatgpt_copycat}
    \vspace{-3mm}
\end{figure}

\end{document}